\documentclass{article}

\usepackage{PRIMEarxiv}
\usepackage[spanish]{babel}
\usepackage[utf8]{inputenc} 
\usepackage[T1]{fontenc}    
\usepackage{hyperref}       
\usepackage{url}            
\usepackage{booktabs}       
\usepackage{amsfonts}       
\usepackage{nicefrac}       
\usepackage{microtype}      
\usepackage{lipsum}
\usepackage{fancyhdr}       
\usepackage{graphicx}       
\graphicspath{{media/}}     
\usepackage{array}
\usepackage[sorting=none]{biblatex}
\usepackage{csquotes}

\title{Delimitación de ceniza volcánica mediante Inteligencia Artificial basada en Pix2Pix}

\author{
  Christian Carrillo, Gissela Torres, Christian Mejía-Escobar \\
  Carrera de Geología, FIGEMPA \\
  Universidad Central del Ecuador \\
  Quito, Ecuador\\
  \texttt{\{cfcarrillo, getorresg, cimejia\}@uce.edu.ec} \\
}

\begin{document}
\maketitle

\begin{abstract}
Las erupciones volcánicas emiten ceniza que puede ser dañina para la salud humana y causar afectaciones a la infraestructura, las actividades económicas y el ambiente. La delimitación de nubes de ceniza permite conocer su comportamiento y dispersión, lo cual ayuda en la prevención y mitigación de este fenómeno. Los métodos tradicionales aprovechan programas de software especializados para procesar las bandas o canales que componen las imágenes satelitales. Sin embargo, su uso se limita a expertos, demanda mucho tiempo y recursos computacionales significativos. En los últimos años, la Inteligencia Artificial ha marcado un hito en el tratamiento computacional de problemas complejos en diferentes áreas. En especial, las técnicas de Deep Learning permiten un procesamiento automático, ágil y preciso de las imágenes digitales. El presente trabajo propone el uso del modelo Pix2Pix, un tipo de red generativa adversaria que, una vez entrenada, aprende el mapeo de imágenes de entrada a imágenes de salida. La arquitectura de dicha red compuesta por un generador y un discriminador ofrece la versatilidad necesaria para producir imágenes de nubes de ceniza en blanco y negro a partir de imágenes satelitales multiespectrales. La evaluación del modelo, basada en gráficas de pérdida y precisión, una matriz de confusión, y la inspección visual, indica una solución satisfactoria para la delimitación precisa de nubes de ceniza, aplicable en cualquier zona del mundo y se convierte en una herramienta útil en la gestión de riesgos.
\end{abstract}


\section{Introducción}
\label{intro}
Los volcanes se encuentran en casi todo el mundo, sobre la superficie terrestre y bajo el mar, formados como grandes montañas o pequeñas colinas, algunos activos y otros que han pasado largos períodos de tiempo en reposo. En el caso de una erupción volcánica, producen fracturas en la superficie terrestre por las cuales asciende material rocoso fundido llamado magma emitiendo consigo gases tóxicos y ceniza \cite{Rivera}\cite{IGN1}. La ceniza es un material compuesto por roca muy fina que produce afectaciones económicas y materiales en el ecosistema, la industria agrícola y ganadera, las infraestructuras, los servicios básicos, así como daños en la salud humana, por ejemplo, enfermedades respiratorias, quemaduras y lesiones oculares\cite{Rivera}\cite{IGN3}.

Conocer el comportamiento de la dispersión de la nube de ceniza ayuda en la prevención y mitigación de este fenómeno, lo cual es de importancia fundamental en la gestión de riesgos \cite{IGN3}. Mediante la localización y delimitación de una nube de ceniza volcánica, es posible determinar la distancia de propagación con respecto a las ciudades cercanas al volcán, prever y mitigar el impacto a las personas en el área, evitando consecuencias como enfermedades, pérdidas económicas y afectaciones en el ecosistema \cite{IGN3}. De manera general, dicha localización y delimitación es realizada por expertos en el área a través de programas de software especializados, mismos que procesan las imágenes de satélite (multiespectrales) combinando bandas, pero ocupando mucho tiempo, esfuerzo y recursos significativos \cite{Velastegui}, por lo que se ve la necesidad de buscar un método alternativo que permita agilitar el proceso de manera automática y en tiempo real.

En la actualidad, la Inteligencia Artificial (IA) se ha convertido en una alternativa exitosa para enfrentar problemáticas de muchas áreas. Específicamente, las técnicas de Deep Learning, un subcampo de la IA, han servido para el procesamiento automático de imágenes y abordar problemas como clasificación, identificación, localización, y segmentación de objetos \cite{Abeluk}. La delimitación de nubes de ceniza volcánica en una imagen satelital se enmarca en una problemática de segmentación de objetos, generalmente tratada con Redes Neuronales Convolucionales (CNNs) dispuestas en una arquitectura de tipo encoder-decoder (codificador-decodificador) \cite{Carrion}. Como precedente tenemos el trabajo "Delimitación de ceniza volcánica en imágenes satelitales mediante Deep Learning" \cite{Aldas}, el cual aprovecha una estructura de red convolucional como codificador y una red deconvolucional como decodificador, para el entrenamiento de un dataset de imágenes satelitales multiespectrales y sus pares correspondientes de imágenes de nubes de ceniza segmentadas en blanco y negro. Así es posible la generación de la imagen con la delimitación de ceniza volcánica tomando el Volcán Sangay como caso de estudio.

\begin{figure}[!htb]
        \centering
        \includegraphics[width=\textwidth]{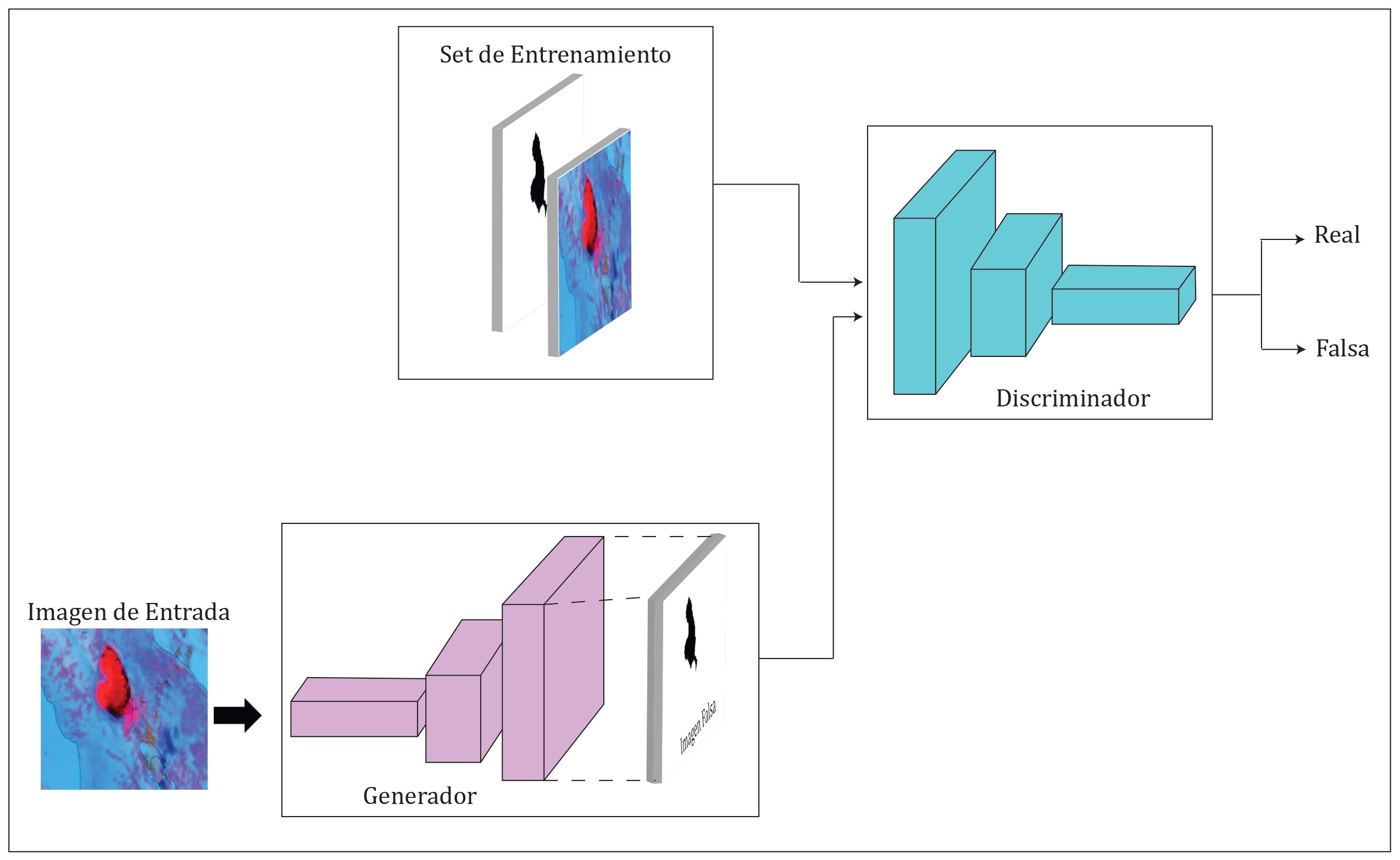}
        \caption{Esquema representativo de una red GAN.}
        \label{fig:0}
    \end{figure}

Nuestro trabajo propone un nuevo enfoque basado en una red neuronal de tipo generativa adversaria (GAN). Es una manera nueva de utilizar deep learning para la generación de imágenes con la aplicación de dos redes neuronales profundas que son el \textit{generador}, encargado de presentar las muestras que queremos crear, mientras que el \textit{discriminador} analiza si el producto generado por la red pertenece o no al conjunto de entrenamiento \cite{Goofellow}. Dicha arquitectura es representada en la Figura \ref{fig:0}.

Una de las implementaciones más populares de las GANs es el modelo \textit{Pix2Pix}. Este modelo es un algoritmo a partir de el cual se construye y entrena una red antagónica generativa condicional, el objetivo es realizar un mapeo de una imagen de entrada a una imagen de salida (traducción). Este mapeo es secuencial y se lo realiza de imagen a imagen. Pix2Pix se puede adaptar y aplicar a varias tareas, por ejemplo, síntesis de fotos desde mapas de etiquetas, generación de imágenes de fotos coloreadas a partir de imágenes en blanco y negro, conversión de fotos de Google Maps en imágenes aéreas, transformar bocetos en fotos, etc. \cite{Isola}. Uno de los desarrolladores de este algoritmo es Phillip Isola de la Universidad de California que desde el año 2008 viene realizando trabajos relacionados a IA. En 2016 revolucionó el campo de las GANs con su trabajo “Image-to-Image Translation with Conditional Adversarial Networks” usando el modelo Pix2Pix. Este trabajo no solo se enfocó en la traducción de imagen de entrada a imagen de salida, también especificó las funciones de pérdida al evaluar el modelo. Otros desarrolladores destacados en GANs son Jun-Yan Zhu de la Universidad de Carnegie Mellon y Thinghui Zhou de la Universidad de California que comparten publicaciones con Phillip Isola \cite{Isola}. En la actualidad, se encuentran varias publicaciones asociadas a GANs, específicamente al modelo Pix2Pix, muchas están disponibles en plataformas de GitHub, OneDrive, Machine Learning Mastery, IArtificial.net, keras.io, Google Colab, etc., todas en lenguaje de programación Python.

Nuestra hipótesis establece que si una red GAN es capaz de generar imágenes que son muy similares a las de un conjunto de entrenamiento, y por lo cual es un modelo bastante utilizado para el aumento artificial de datos (\textit{data augmentation}), también puede ser capaz de generar segmentaciones de objetos en imágenes si el dataset de entrada está constituído por este tipo de imágenes. De esta forma, es posible dar solución a un problema geológico basado en GANs que permita delimitar la nube de ceniza volcánica en una zona específica a partir de imágenes satelitales multiespectrales e imágenes en blanco y negro mediante la aplicación del modelo \textit{Pix2Pix}.

La implementación del modelo tendrá un alto impacto en varios sectores, tal como la gestión de riesgos \cite{IGN3}. La manera de procesar la información relacionada con un evento de erupción y el tiempo que ocupa cambiará drásticamente. La capacidad de realizar automáticamente la delimitación de la nube de ceniza usando imágenes satelitales multiespectrales en tiempo real permite la identificación de zonas con potencial afectación (zonas agrícolas, poblados cercanos o infraestructura) antes, durante y después de una erupción volcánica, gestionando la crisis y los recursos adecuadamente. Las contribuciones concretas de nuestro estudio son:

\begin{itemize}
    \item Aplicación de un modelo GAN para resolver un problema de segmentación de objetos en una imagen, específicamente, la delimitación de nubes de ceniza volcánica en imágenes satelitales.
    
    \item Elaboración de un dataset apto para GAN a partir del utilizado en \cite{Aldas}, por medio de un redimensionamiento y unión de las imágenes de entrada y salida.
    
    \item Generación del código de Pix2Pix en una estructura de clases, que permite tener mejor legibilidad, organización, comprensión y facilidad de implementación. 
\end{itemize}

El código reorganizado en clases del modelo Pix2Pix, el repositorio de los pares de imágenes satelitales multiespectrales e imágenes en blanco y negro de nube de ceniza, las imágenes de prueba para la predicción del modelo, así como los resultados obtenidos durante el entrenamiento y predicción del modelo son de acceso público en \textit{GitHub}\footnote{\url{https://github.com/GisseTorres04/GANs\_Pix2Pix\_Traduccion\_Imagenes\_Ceniza.git}}. Los modelos obtenidos durante el entrenamiento para aplicar en la predicción se encuentran disponibles en \textit{Google Drive}\footnote{\url{https://drive.google.com/drive/folders/1APEPJjxSkJ0Ev5f6W2QS8NlbskSchgl2?usp=sharing}}.

El resto del documento se estructura de la siguiente manera: la sección \ref{state-of-art} cita algunos trabajos relacionados; la sección \ref{methodology} detalla la metodología empleada para el desarrollo del proyecto; la sección \ref{experiments} describe el entrenamiento, evaluación y predicción del modelo; la sección \ref{discusion} presenta una tabla comparativa de los modelos CNNs y GANs, con sus ventajas y desventajas; por último, la sección \ref{conclusion} enuncia las conclusiones y recomendaciones acerca de posibles trabajos futuros.

\section{Trabajos relacionados}
\label{state-of-art}
Según nuestro conocimiento, no existen implementaciones de GANs y, en particular, de Pix2Pix, aplicadas a la delimitación de ceniza volcánica. Sin embargo, es muy conocida la \textit{Pix2Pix GAN for Image-to-Image Translation}, donde se produce la traducción de imágenes satelitales a imágenes de Google Maps con el fin de poder obtener una imagen en plano a partir de una imagen que contiene varios objetos, cuya aplicación extrae los rasgos más relevantes de la imagen satelital. En este caso fue para el reconocimiento y generación de vías, zonas de vegetación y cuerpos de agua \cite{Joyce}. También se hace la traducción de imágenes a partir de imágenes de objetos reales y bocetos a color o en blanco y negro de los mismos, la imagen de salida corresponde a una predicción del boceto de entrada, por ejemplo, se ingresa una imagen en boceto en blanco y negro de una cartera, mientras el modelo genera una imagen de una cartera a color con relieve, generando una imagen con una proyección realista \cite{Isola}. Nuestro interés se enfoca en el único trabajo que aborda nuestra misma temática, presentado por Aldás et. al \cite{Aldas}, el cual pretendemos extender y obtener mejores resultados a través de una arquitectura GAN.

A partir del análisis de resultados del citado estudio, y la incertidumbre geológica que hace referencia a la dispersión de la nube de ceniza en la zona de afectación, este patrón de dispersión se asocia a un fenómeno de tipo dinámico, lo cual sugiere que tendremos una delimitación de ceniza de acuerdo a una escala cronológica de tiempo, es decir, la delimitación de la nube de ceniza depende del momento exacto en que fue tomada la imagen satelital multiespectral. Teniendo en cuenta estas consideraciones y con base en los resultados obtenidos con una CNN, surge la necesidad de buscar un modelo que presente mejores resultados, para lo cual proponemos la aplicación de redes generativas adversarias (GANs), a partir del desarrollo de un modelo Pix2Pix, el cual consiste en identificar zonas de nube de ceniza mediante el mapeo de imágenes satelitales multiespectrales.

\section{Materiales y métodos}
\label{methodology}

\begin{figure}[!htb]
        \centering
        \includegraphics[width=\textwidth]{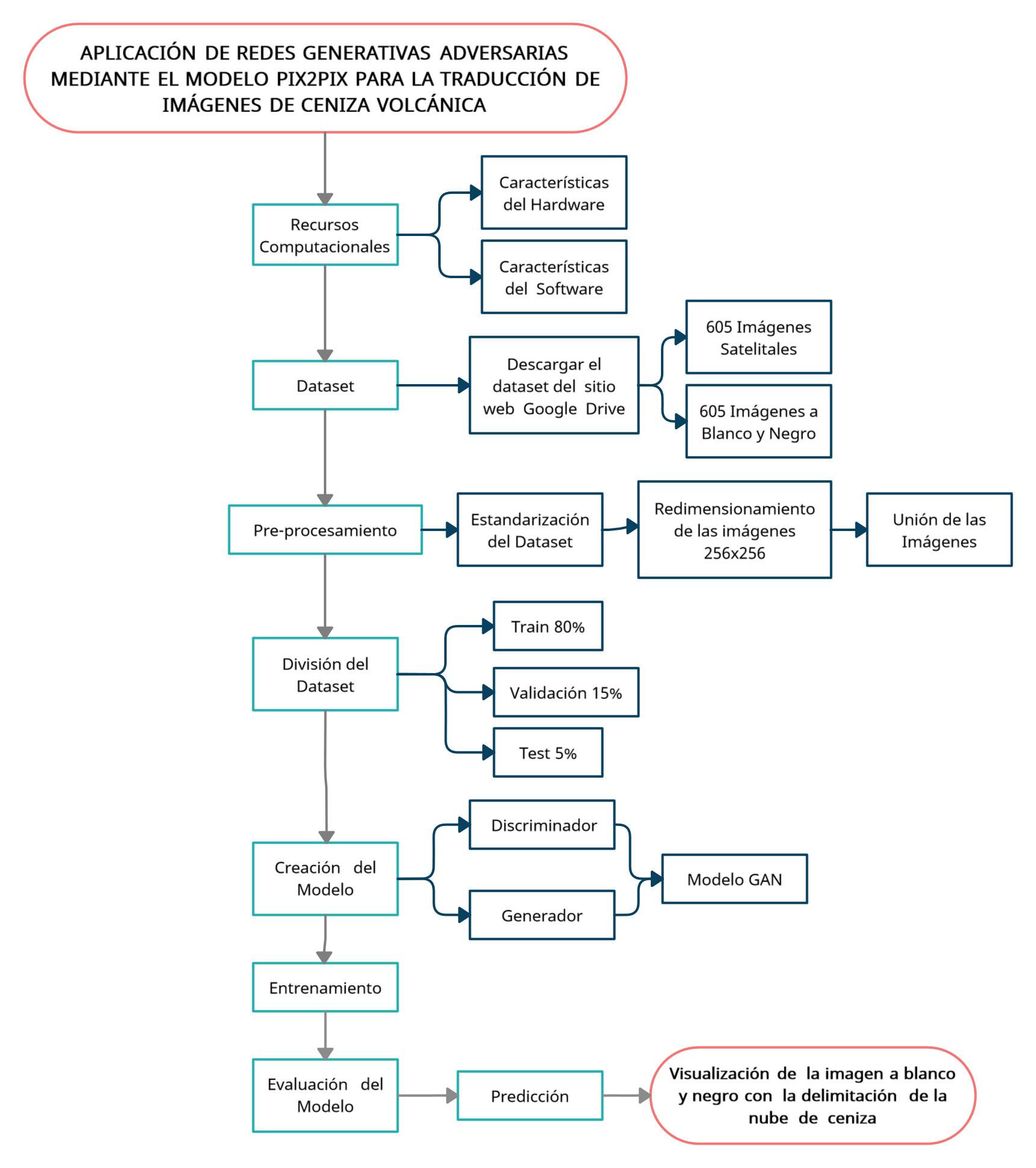}
        \caption{Flujograma de la metodología del trabajo desarrollado.}
        \label{fig:workflow}
    \end{figure}

El propósito es identificar y delimitar una nube de ceniza volcánica en imágenes satelitales mediante una solución basada en aprendizaje automático supervisado. A partir de un dataset de pares de imágenes (multiespectrales y sus correspondientes imágenes en blanco y negro), se entrena un modelo de Deep Learning de tipo GAN, que posteriormente servirá para la delimitación de nube de ceniza volcánica en nuevas imágenes satelitales. Seguidamente, la Figura \ref{fig:workflow} presenta el flujo de trabajo para el desarrollo del proyecto y, a continuación, se describe en detalle cada una de las etapas representadas en el diagrama de flujo.

\subsection{Dataset}
\begin{figure}[!htb]
        \centering
        \includegraphics[width=\textwidth]{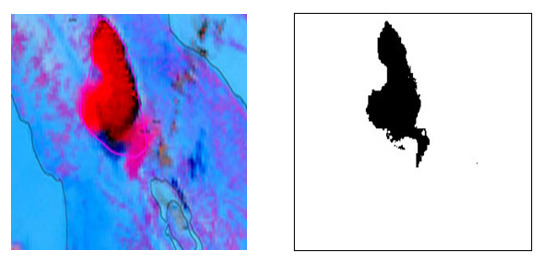}
        \caption{A la izquierda se puede observar la imagen satelital multiespectral, mientras que a la derecha se observa la imagen en blanco y negro representando la nube de ceniza.}
        \label{fig:dataset}
    \end{figure}

El insumo principal para un modelo de aprendizaje automático son los datos de entrada con su respectiva salida o respuesta. En nuestro caso, imágenes satelitales con presencia de nube de ceniza volcánica y su correspondiente delimitación en imágenes de blanco y negro. Hemos aprovechado el dataset utilizado por Aldás et. al \cite{Aldas}, disponible públicamente en \textit{Google Drive}\footnote{\url{https://drive.google.com/drive/folders/1jFAg5MiF9zZK1ESlhGP7usHTs4zeprGU}}. Consiste de 2 carpetas, una para las 600 imágenes satelitales multiespectrales y otra para las 600 imágenes en blanco y negro. El formato de todas las imágenes es JPG, representando volcanes alrededor del mundo y ocupan un peso de 25.8 MB. Un ejemplo se muestra en la Figura \ref{fig:dataset}.

\subsection{Preprocesamiento}
\begin{figure}[!htb]
        \centering
        \includegraphics[width=\textwidth]{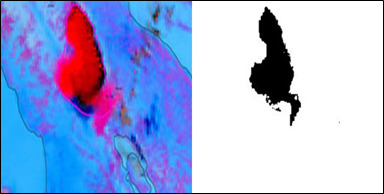} 
        \caption{Unión de imágenes multiespectrales e imagen en blanco y negro con un tamaño de 256x512 px.}
        \label{fig:union}
    \end{figure}

En esta etapa realizamos dos modificaciones a las imágenes del dataset original. En primer lugar, con la herramienta en línea de la página Web de \textit{PineTools}\footnote{\url{https://pinetools.com/es/redimensionar-imagen-lotes-serie}}, redimensionamos el tamaño de las imágenes satelitales multiespectrales de nube de ceniza e imágenes en blanco y negro a 256x256 píxeles. Este tamaño permite tener una estandarización y un reconocimiento de los píxeles de manera que pueda obtener un resultado más proximo al real. En segundo lugar, llevamos a cabo la unión de los pares de imágenes como una sola de manera manual con la ayuda de la herramienta de combinar imágenes en el sitio Web \textit{PineTools}\footnote{\url{https://pinetools.com/es/combinar-imagenes}}, obteniendo un tamaño de imagen de 256x512 píxeles (Figura \ref{fig:union}).   Estas imágenes son la entrada para la parte discriminadora del modelo Pix2Pix, y que luego será contrastada con la imagen proporcionada por el generador.

Seguidamente, se realiza una normalización de los valores de los píxeles de las nuevas imágenes con el fin de tener un mejor rendimiento del modelo GAN \cite{Microsoft1}. Un píxel tiene un rango entre 0 y 255, donde 0 representa negro y 255 blanco \cite{Calcagni}, luego de la normalización sus valores se escalan a un menor rango entre -1 y 1, lo que permite optimizar el proceso de entrenamiento.

\subsection{División del dataset}         
Una tarea fundamental del aprendizaje automático es la subdivisión del conjunto de datos en tres: \textit{entrenamiento} (train) es el subconjunto utilizado para ajustar el modelo; \textit{validación} (val) con el objetivo de proporcionar una evaluación imparcial del ajuste de un modelo en el conjunto de datos de entrenamiento mientras se selecciona el mejor modelo y \textit{prueba} (test) para proporcionar una evaluación imparcial del ajuste del modelo final \cite{Towards}. La  división de  las  imágenes  es  realizada  de  manera  aleatoria basándose  en la  cantidad de imágenes satelitales según el tipo de satelite: GOES 16 y 17, Meteosat-11 y Himawari-8. Esta división se realiza de acuerdo con el porcentaje proporcionado para train, val y test. La Tabla \ref{tab1} muestra la cantidad de imágenes por cada tipo de satélite, así como la cantidad que contiene cada conjunto de datos.

    \begin{table}[!htb]
        \centering
        \caption{División del dataset de imágenes satelitales.}\label{tab1}
        \vspace{0.2cm}
        \begin{tabular}{|c|c|c|c|c|}
        \hline
        \textbf{TIPO DE SATÉLITE} &  \textbf{N°} & \textbf{Train} & \textbf{Val} & \textbf{Test}\\
        \hline
        \textbf{GOES16}  & 148 & 118 & 23 & 7\\
        \hline
        \textbf{GOES17} & 401 & 320 & 61 & 20\\
        \hline
        \textbf{Himawari-8} & 16 & 12 & 3 & 1\\
        \hline
        \textbf{Meteosat-11 }& 35 & 28 & 5 & 2\\
        \hline
        \textbf{TOTAL}  & 600 & 478 & 92 & 30\\
        \hline
        \textbf{PORCENTAJE} &  \textbf{100\%} & \textbf{80\%} & \textbf{15\%} & \textbf{5\%}\\
        \hline
        \end{tabular}
    \end{table}

Como resultado, la estructura del dataset se encuentra dividido en tres carpetas: \textit{train} que contiene 478 (80\%) pares de imágenes, mientras la carpeta \textit{val} contiene 92 (15\%) pares de imágenes y el \textit{test} 30 (5\%) pares de imágenes. Esta división se debe a que se desea que el modelo tenga la cantidad suficiente de ejemplos para su entrenamiento y no caiga en un sobreajuste (\textit{overfitting}).

\subsection{Modelo}
El modelo Pix2Pix permite una traducción de una imagen fuente a una de destino, para lo cual incorpora una parte discriminadora y una generadora, que compiten entre sí, de manera que la primera trata de distinguir y la segunda mejora en la generación de imágenes similares a las del dataset de entrada. A continuación, describimos en detalle esta estructura:

\begin{itemize}
    \item \textbf{Discriminador}: es una red neuronal convolucional profunda que realiza la clasificación de imágenes (Figura \ref{fig:discriminador}). En concreto, una clasificación de imágenes condicional \cite{MLM}.
    \begin{itemize}
    
    \item Toma  como entrada la imagen de origen (una imagen satelital multiespectral de nube de ceniza), y la imagen objetivo-traducida (una imagen en blanco y negro que representa la delimitación de la nube de ceniza) que pertenece al dataset, así como también el resultado de la generación de la imagen objetivo por el generador. Ambas imágenes tienen una dimensión $w$x$h$x$c$ de 256x256x3 ($w$=ancho, $h$=altura, $c$=canales).
    
    \item Antes de entrenar, se realiza la superposición de imágenes que se requiere para que el modelo aprenda a mapear los píxeles de la imagen de entrada a los de salida.
    
    \item El  modelo cuenta con  5 capas convolucionales donde se aplican 64, 128, 256, 512, y 512 filtros \textit{kernels}, respectivamente. En cada capa convolucional se aplican los siguientes hiperparámetros: \textit{tamaño de filtro} 4x4  píxeles, \textit{strides} de 2x2 y un \textit{padding} de tipo ''same'', intercaladas con una capa de \textit{batch normalization} que permite normalizar cada lote de los datos permitiendo que la red neuronal trabaje de mejor manera \cite{Ioffe}. A cada  capa de convolución le  sigue  una  función  de  activación  de  tipo  LeakyReLU  penalizando  los  valores negativos  mediante  el  coeficiente  rectificador y convirtiéndolos a valores positivos \cite{Antona}. Se crea un número de \textit{mapas de características} que capturan patrones de la imagen, determinado por el número de filtros aplicados. 
    
    \item El diseño del discriminador se basa en el campo receptivo del modelo, un campo receptivo se define como una región limitada del espacio dentro de una imagen de entrada, que afectan a una característica concreta dentro de la red \cite{Baeldung}, la cual define la relación entre una salida del modelo y el número de píxeles de la imagen de entrada. Este diseño particular de un modelo GAN, se denomina modelo \textit{PatchGAN} y se diseña cuidadosamente para que cada predicción de salida del modelo se corresponda con un cuadrado o patch de 70×70 de la imagen de entrada \cite{MLM}. 
    
    \item  El patch produce como salida un mapa de activación de valores de predicciones (probabilidades) reales o falsas que se puede promediar para obtener una única puntuación (valor de pérdida), en lugar de considerar un output que represente toda la imagen de una sola vez. Cada patch de 16×16 representa a una región de la imagen de 70x70 píxeles.
    
    \item La capa de salida (patch) constituye un \textit{número de filtro} de 1 con un tamaño 4x4, \textit{padding} de tipo ''same'', seguido de una capa de función de activación de tipo \textit{Sigmoid}, esta función permite convertir los valores entre 0 y 1 correspondiente al mapa de activación de valores \cite{Antona}. El mapa de activación de valores constituye un patch con valores únicos de 1, lo que significa que la imagen predicha es la correcta con su entrada, mientras que un patch con valores únicos de 0, indican que la imagen predicha no pertenece a la entrada, es decir, cada valor es una probabilidad de que un patch de la imagen de entrada sea real.
\end{itemize}

        \begin{figure}[!htb]
            \centering
            \includegraphics[width=\textwidth]{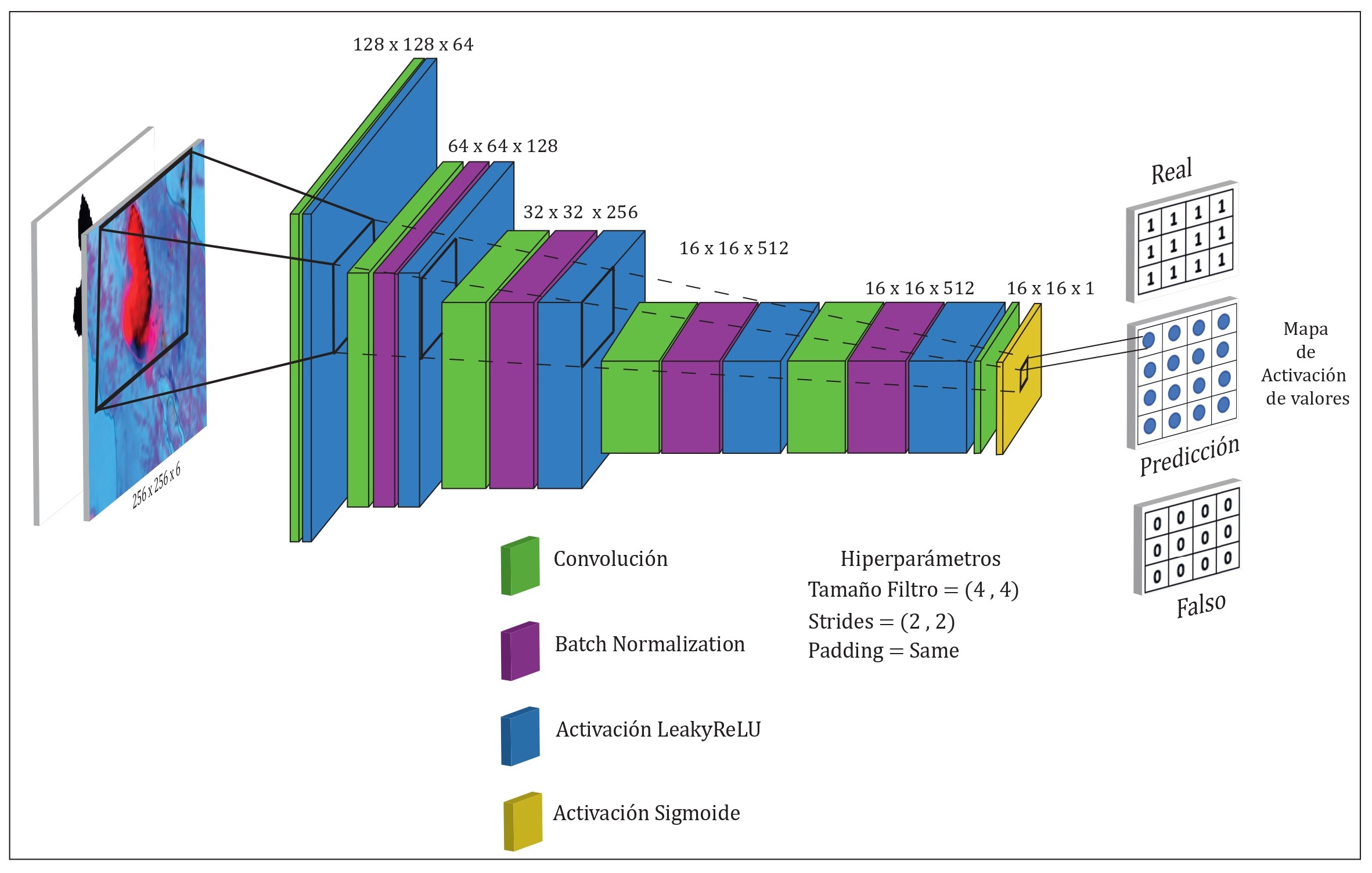} \caption{Arquitectura del discriminador a partir de una red neuronal convolucional.}
            \label{fig:discriminador}
        \end{figure}

    \item \textbf {Generador}: es un modelo codificador-decodificador que utiliza una arquitectura \textit{U-Net} (Figura \ref{fig:generador}). El modelo toma una imagen de origen (una imagen satelital multiespectral de nube de ceniza) y genera una imagen de destino (una imagen en blanco y negro de nube de ceniza). Para ello, primero reduce el muestreo (\textit{downsampling}) o la codificación de la imagen de entrada hasta una capa de cuello de botella, y luego aumenta el muestreo (\textit{upsampling}) o la decodificación de la representación de cuello de botella hasta el tamaño de la imagen de salida. La arquitectura \textit{U-Net} significa que se añaden conexiones de salto entre las capas de codificación y las capas de decodificación correspondientes, formando una ''U'' \cite{MLM}.
    
    \begin{itemize}
        \item La imagen de entrada tiene dimensiones 256x256x3, las dos primeras corresponden al tamaño y la última a los 3 canales de color RGB.        
        \item El encoder (codificador) permite la identificación, extracción de características de la imagen y la reducción de la dimensión de las mismas (\textit{downsampling}) \cite{Carrion}.
        Consta de 7  capas convolucionales,  en donde se aplican 64, 128, 256, 512, 512, 512, y 512 filtros, en cada capa convolucional se aplican los siguientes hiperparámetros: \textit{tamaño de filtro} de 4x4, \textit{stride} de 2x2, y \textit{padding} de tipo ''same'', intercalada con una capa de \textit{batch normalization}. A cada capa de convolución le sigue una función de activación \textit{LeakyReLU}, posteriormente se define el cuello de botella aplicado a la U-Net del generador donde se utiliza un tamaño de filtro de 4x4 con un \textit{stride} de 2x2 aplicando la función \textit{ReLu} que permite anular los valores negativos y deja pasar a los positivos en la capa \cite{Antona}.        
        \item El decoder (decodificador) permite recuperar las dimensiones originales de las imágenes de entrada mediante un muestreo ascendente (\textit{upsampling}) obteniendo las características principales de la imagen \cite{Carrion}. Contiene 7 capas de convolución transpuesta que es un tipo de deconvolución, es decir, funciona de manera inversa a la convolución normal \cite{Calcagni}, que consta de 512, 512, 512, 512, 256, 128, y 64 filtros para cada capa, respectivamente. Con el objetivo de obtener la misma imagen de entrada con sus características mínimas, en unión con los atributos requeridos por el discriminador, además es intercalada con una capa de \textit{batch  normalization}, se añade  la  condición \textit{dropout} que desactiva un número de neuronas de forma aleatoria, ayudando a reducir un posible overfitting \cite{Srivastava}. Se realiza la concatenación permitiendo la conexión entre el encoder y el decoder; por último, se añade la función \textit{ReLU}. En cada capa de convolución transpuesta se aplican los siguientes hiperparámetros: \textit{tamaño de filtro} de 4x4, \textit{stride} de 2x2, y \textit{padding} de tipo ''same''.    
        \item En la capa de salida se aplica una convolución transpuesta con un filtro de 4x4, con \textit{stride} de 2x2, y una función de activación de tipo \textit{Tangente Hiperbólica} (tanh) que permite transformar los valores introducidos a una escala entre -1 y 1 \cite{Antona}.
    \end{itemize}

        \begin{figure} [!htb]
            \centering
            \includegraphics[width=\textwidth]{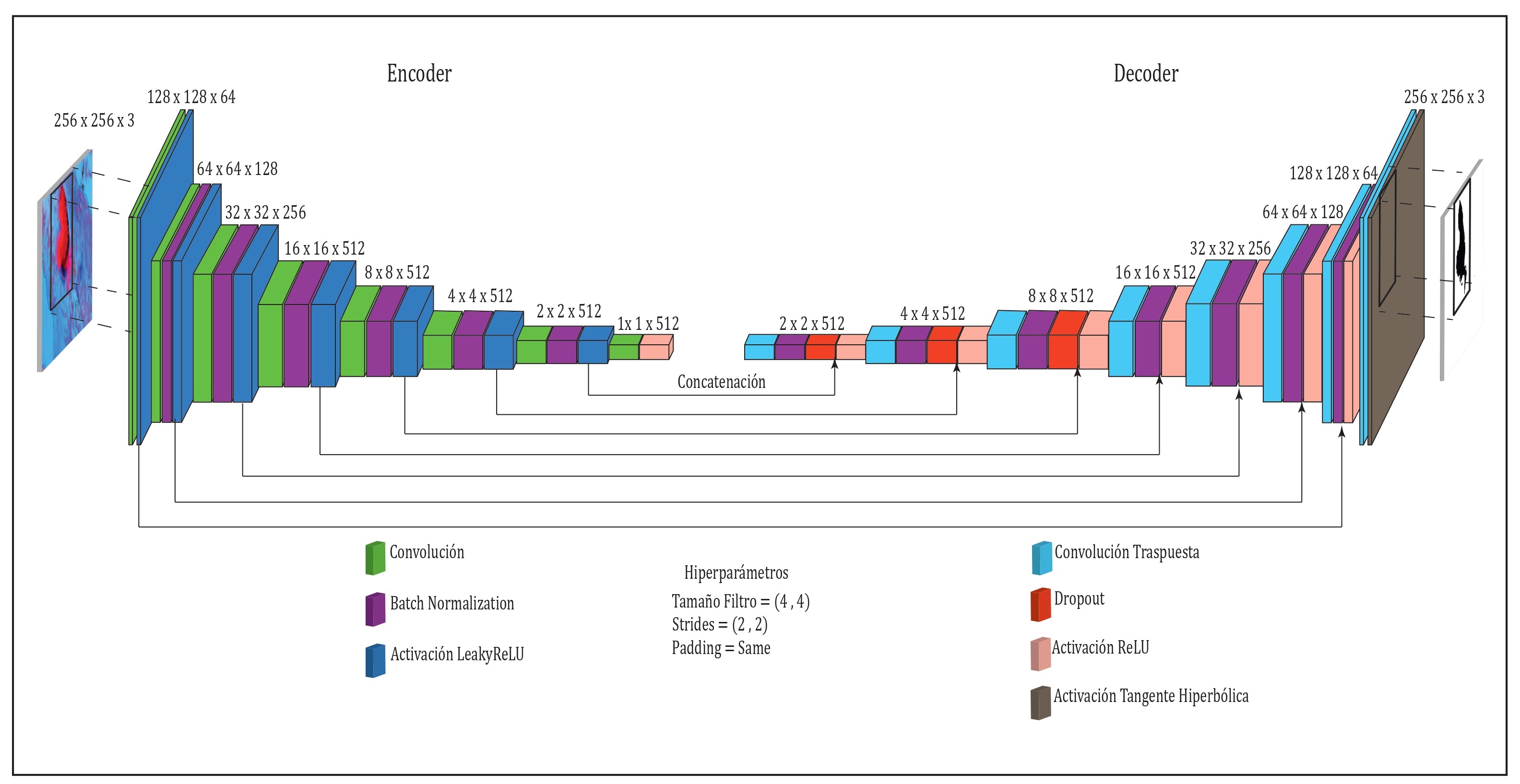} \caption{Arquitectura del generador a partir de una red neuronal convolucional de tipo U-Net.}
            \label{fig:generador}
        \end{figure}

    \item \textbf {Red Generativa Adversaria (GAN)}: toma los modelos generador y discriminador ya definidos como argumentos y utilizando la API funcional de Keras (un modelo es la entrada de otro modelo) \cite{Antona}, los conecta juntos en un modelo compuesto, debido a que la salida del generador se conecta al discriminador como la correspondiente imagen de entrada. Se proporciona una imagen satelital de la nube de ceniza como entrada al generador y al discriminador, también recibe una imagen en blanco y negro de la delimitación de la nube de ceniza para que pueda comparar con la imagen generada. Una vez finalizado el generador, envía la imagen de salida (imagen en blanco y negro de la delimitación de la nube de ceniza) como entrada al discriminador, en donde este predice la probabilidad de que el generador sea una traducción real de la imagen de origen (imagen en blanco y negro), tal como se representa en la Figura \ref{fig:gan}.

        \begin{figure*}[ht]
            \centering
            \includegraphics[width=\textwidth]{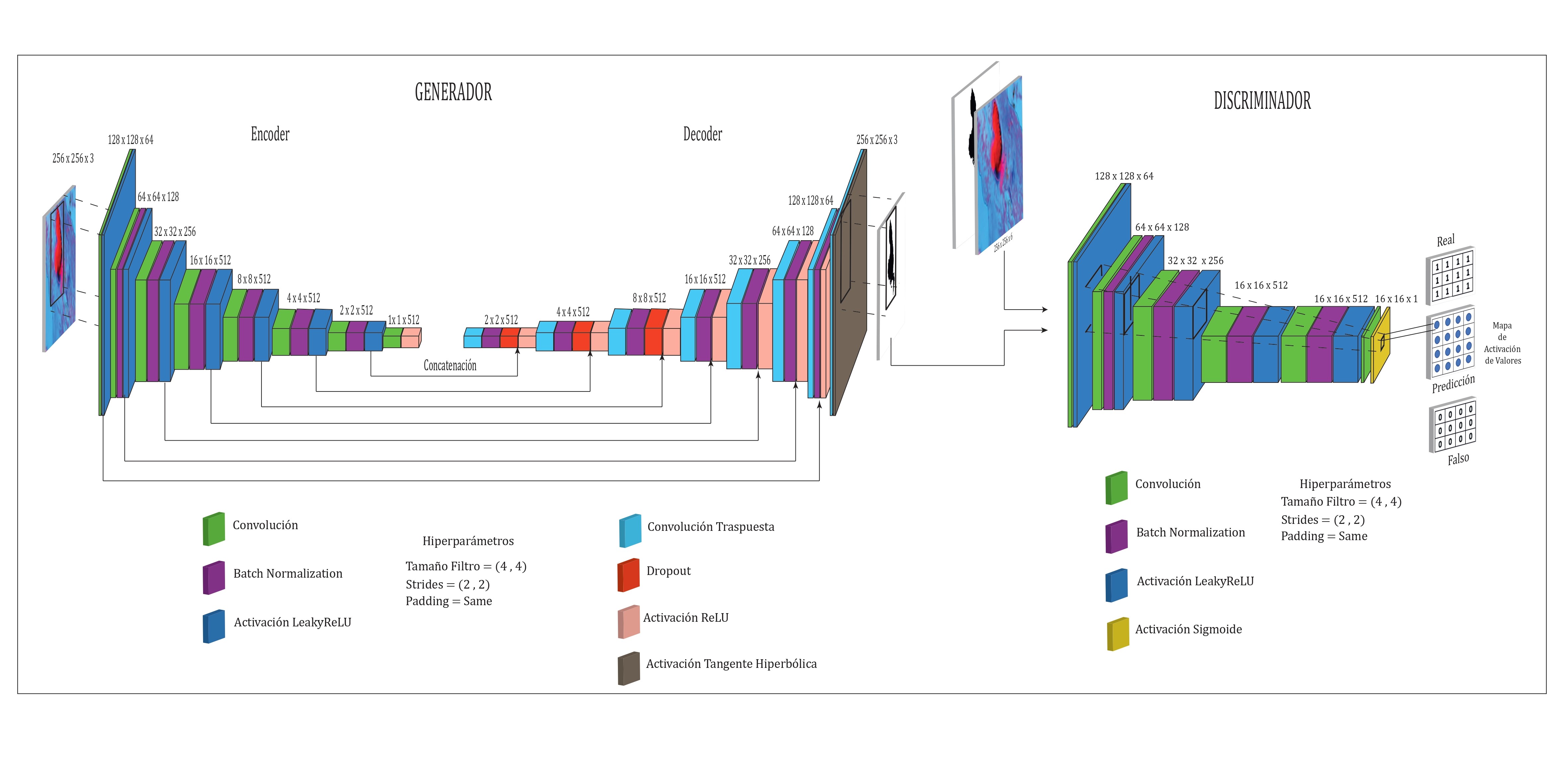} \caption{Arquitectura de la GAN, donde se conectan el modelo generador y el modelo discriminador.}
            \label{fig:gan}
        \end{figure*}
    \end{itemize}

\section{Experimentos y resultados}
\label{experiments}

\subsection{Recursos computacionales}
La plataforma de hardware consiste de una computadora tipo laptop, marca HP modelo Pavion, procesador Intel(R) Core(TM) i7-1065G7, de 1.50 GHz de velocidad, procesador gráfico integrado Intel Iris Plus, 16 GB de RAM, disco duro SSD de 1TB de almacenamiento. Para el desarrollo del proyecto se dispone del sistema operativo Windows 10 Home \cite{Microsoft}, lenguaje de programación Python versión 3.9.7 \cite{Python} conjuntamente con el entorno de desarrollo integrado (IDE) Visual Studio Code \cite{Visual}. Además, las librerías de programación generales \textit{OS}, para crear carpetas y manipular los archivos \cite{Covantec}, así como \textit{NumPy} para la parte matemática, y \textit{Matplotlib} empleada en la generación de gráficas \cite{Stack}. En cuanto a librerías específicas para el aprendizaje automático, tenemos \textit{Tensorflow} y \textit{Keras} \cite{Tensor}\cite{Antona}.

\subsection{Entrenamiento del modelo}

Es un proceso iterativo que se realiza con el conjunto de datos de entrenamiento (train) que permite ajustar el modelo, el cual ve y aprende de estos datos \cite{Towards}. Para el entrenamiento del modelo se han tomado las siguientes consideraciones:

\begin{itemize}
    \item El modelo discriminador se entrena directamente con imágenes reales y generadas. En cambio, el modelo generador se entrena a través del modelo discriminador. Se actualizan los pesos para minimizar la pérdida predicha por el discriminador para las imágenes generadas marcadas como reales. De este modo, se dispone a generar más imágenes reales. 
    
    \item Una época se define como todas las iteraciones de entrenamiento del modelo para todas las imágenes del dataset de entrenamiento, es decir, 478 pares de imágenes.
    
    \item El generador actualiza sus pesos en cada iteración de una época para minimizar la pérdida o el error medio absoluto entre la imagen generada y la imagen de entrada.
    
    \item El discriminador se actualiza de forma independiente, por lo que los pesos se reutilizan en este modelo compuesto, pero se marcan como no entrenables. El modelo compuesto se actualiza con dos objetivos, uno que indica que las imágenes generadas eran reales (pérdida de entropía cruzada), forzando grandes actualizaciones de pesos en el generador hacia la producción de imágenes más realistas, y la traducción real ejecutada de la imagen, que se compara con la salida del modelo del generador.
    
    \item Para el entrenamiento del modelo discriminador se define un optimizador de tipo \textit{Adam} que permite actualizar los valores de los parámetros para reducir el error cometido por la red; la \textit{tasa de aprendizaje} de $0.0002$, que es la longitud del paso que da cada vez que decide cambiar de posición; la \textit{función de pérdida} de tipo \textit{binary\_crossentropy}, que es la medida de la distancia entre distribuciones de probabilidad basado en una clasificación binaria \cite{Calcagni} y una \textit{métrica} de tipo \textit{accuracy} que se utiliza para supervisar y medir el rendimiento del modelo \cite{Carrion}.  
\end{itemize}

\begin{figure}[!htb]
        \centering
        \includegraphics[width=\textwidth]{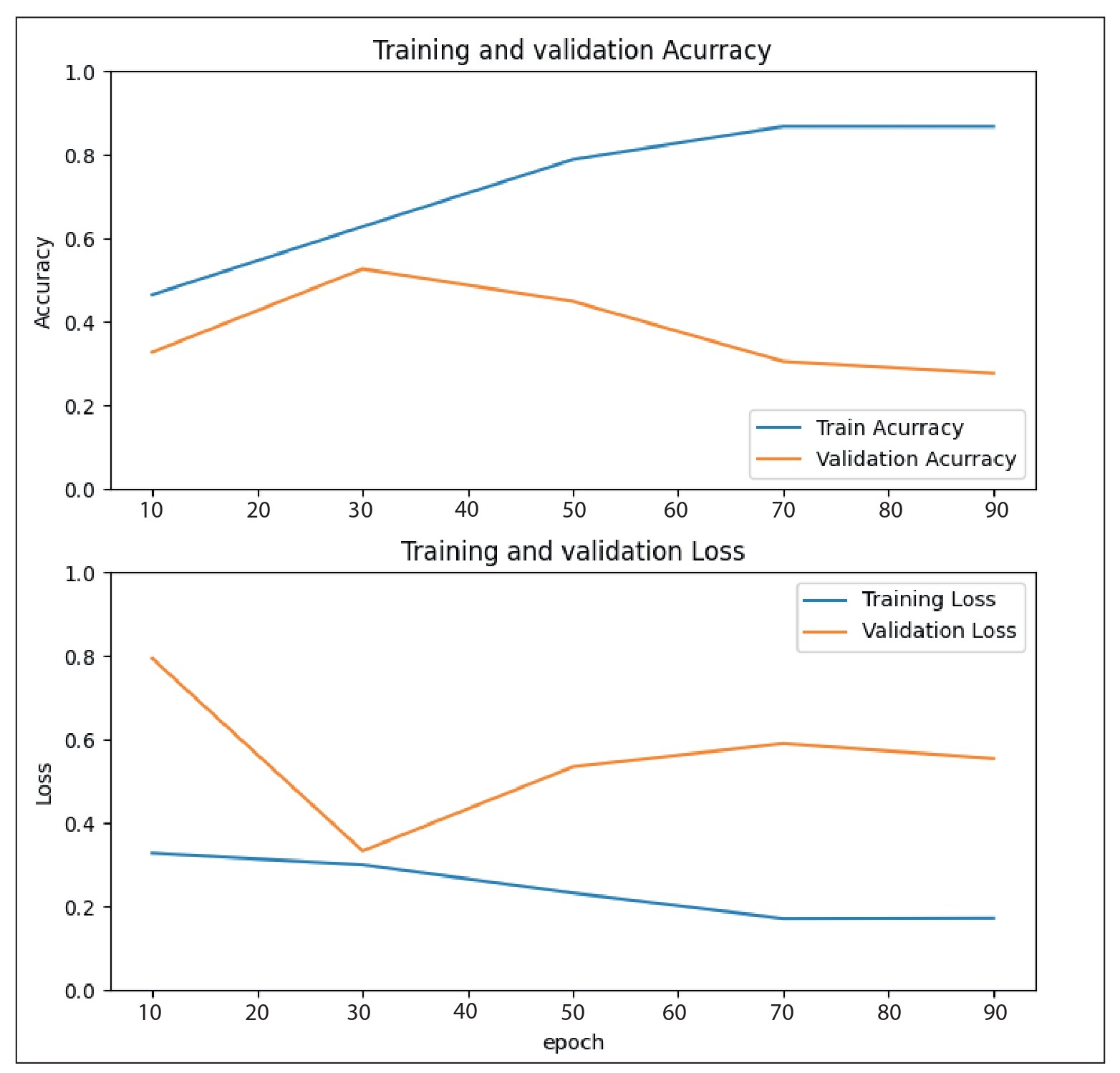} \caption{Curva de precisión durante el entrenamiento (arriba), el valor más alto en la época 30 con 60\%, mientras que la gráfica inferior representa la pérdida durante el entrenamiento, el valor del 37\% en la época 30.}
        \label{fig:loss_acc}
    \end{figure}

El proceso de entrenamiento de la red GAN ocurre de la siguiente manera:
\begin{itemize}

    \item Para el entrenamiento del modelo se define previamente un optimizador \textit{Adam}; la \textit{tasa de aprendizaje} de $0.0002$ y la función de pérdida \textit{binary\_crossentropy}. El número de \textit{épocas} del dataset es de 100 para mantener los tiempos de entrenamiento bajos y se utiliza un \textit{batch\_size} o tamaño de lote de 1.
    
    \item Para cada época de entrenamiento se selecciona primero un lote de imágenes del dataset al azar; en el modelo GAN el lote siempre será de una imagen, por lo que el lote se utiliza para generar primero ejemplos reales y luego se utiliza el generador para producir un lote de muestras falsas, que permite que coincidan con las dimensiones de imágenes de origen reales. A continuación, el discriminador se actualiza con el lote de imágenes reales y luego con las imágenes falsas.
    
    \item El discriminador se entrena mediante la función de Keras (\textit{train\_on\_batch})  que  permite  actualizar  los  pesos  del modelo, además de retornar la pérdida y precisión en el entrenamiento \cite{Microsoft1}.
    
    \item Se  actualiza  el  modelo  del  generador  proporcionando  las  imágenes  reales  de origen como entrada y proporcionando las etiquetas, correspondiente al mapa de activación de valores (1: real, 0: falsa), y las imágenes reales de destino  como  las  salidas  esperadas  del  modelo  necesarias  para  calcular  la  pérdida.  
    
    \item En  cada  iteración  de  entrenamiento,  se obtiene  la  métrica de pérdida  del  discriminador  en  los  ejemplos reales (\textit{d\_loss\_train\_real}),  la  pérdida  del  discriminador  en  los  ejemplos  generados o falsos (\textit{d\_loss\_train\_fake}), adicionalmente se obtiene la precisión del discriminador en  los  ejemplos reales  (\textit{d\_acc\_train\_real}) y  la precisión del  discriminador  en  los  ejemplos  generados  o falsos (\textit{d\_acc\_train\_fake}).
    
    \item El generador  se guarda y el discriminador se evalúa cada 10 épocas de entrenamiento mediante la función de Keras (\textit{evaluate}) en donde se obtienen las métricas de perdida (\textit{d\_loss\_val\_real}), y precisión (\textit{d\_acc\_val\_real}) de la validación del modelo para imágenes reales.
    
    \item La pérdida para cada época de interés se reporta en la consola en cada iteración de entrenamiento y validación del modelo. El historial de pérdidas y precisiones se muestra en la Figura \ref{fig:loss_acc}.
    
\end{itemize}

\begin{figure}[!htb]
        \centering
        \includegraphics[width=\textwidth]{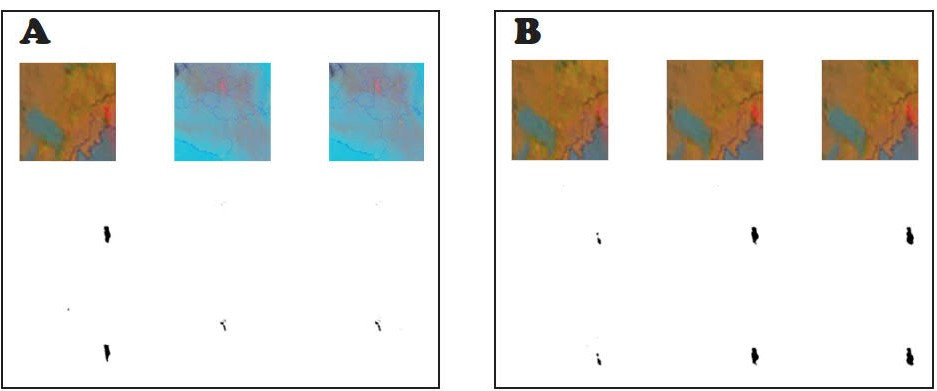} \caption{Comparación de los resultados de la predicción de 2 modelos de distintas épocas. A) Época 10; B) Época 90.}
        \label{fig:10}
    \end{figure}

La red GAN durante el entrenamiento guarda 10 modelos en formato \textit{h5} cada 10 épocas, lo que permite evaluar posteriormente cada uno, eligiendo el modelo que mejor calidad, resolución y ajuste presente para la predicción. Un ejemplo de esta visualización se observa en la Figura \ref{fig:10}, en donde se ilustra la comparación de las salidas de dos modelos en diferentes épocas. En la parte superior e inferior aparecen las imágenes de entrada, mientras que en la parte del medio se observa la imagen generada durante el entrenamiento, obteniendo mejor visualización de la delimitación de la ceniza en la época 90.

\begin{figure}[!htb]
        \centering
        \includegraphics[width=\textwidth]{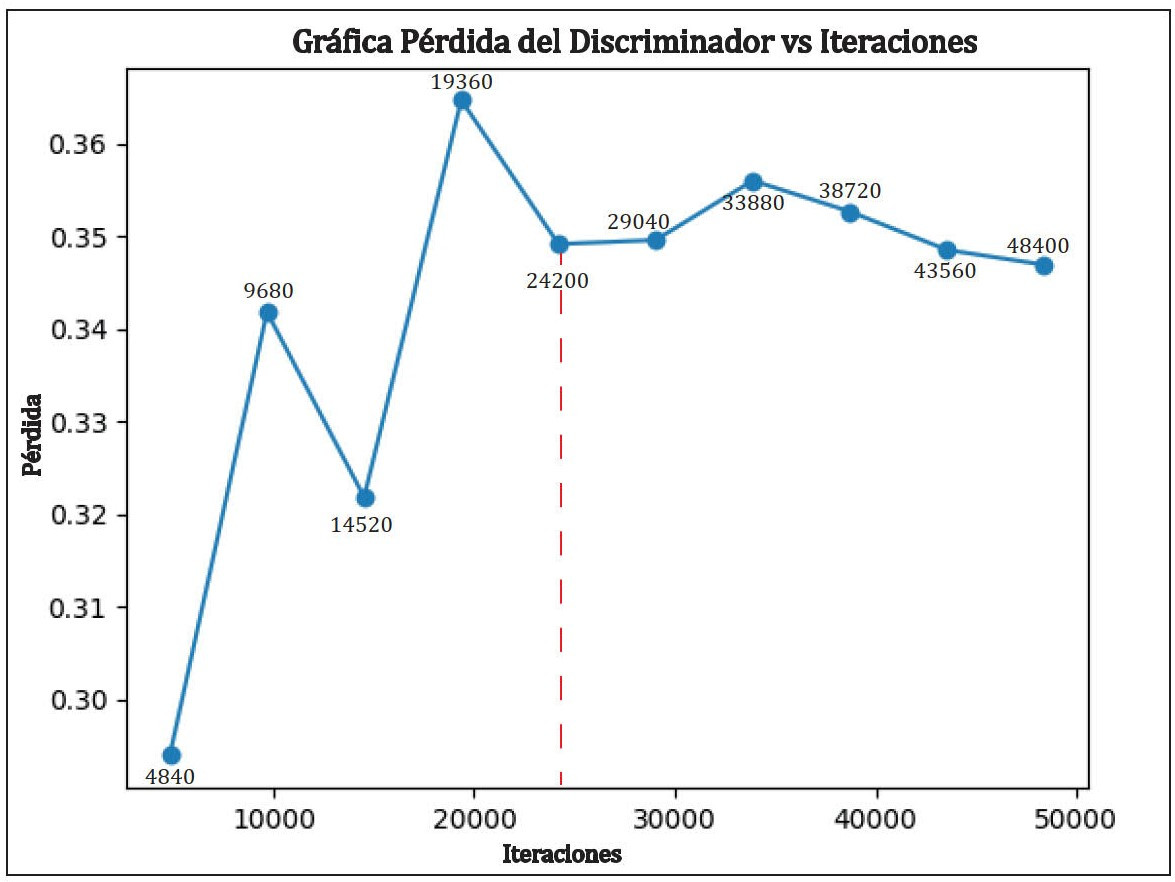} \caption{Pérdida del discriminador para las imágenes generadas o falsas frente al número de iteraciones. La línea roja discontinua separa dos zonas, la del lado izquierdo representa la inestabilidad en la pérdida, mientras que la zona del lado derecho representa la estabilidad en la pérdida.}
        \label{fig:curvaloss}
    \end{figure}

Las imágenes generadas después de la época 50 de entrenamiento empiezan a tener un aspecto muy realista, y la calidad parece seguir siendo buena durante el resto del proceso de entrenamiento (Figura \ref{fig:curvaloss}).

\subsection{Evaluación del Modelo}
Normalmente, los modelos GAN no convergen, sino que se encuentra un equilibrio entre los modelos generador y discriminador, debido a que es del tipo estocástico, es decir, tiene un comportamiento no determinístico, por lo que no se puede saber en qué época se tiene el mejor modelo; a la par no se puede asegurar que conforme aumenten las épocas el modelo será más preciso y a la vez no se puede juzgar fácilmente cuándo debe detenerse el entrenamiento \cite{MLM}. Con el fin de resolver este inconveniente, se  guarda un posible mejor modelo periódicamente durante el entrenamiento cada 10 épocas. Una vez terminado el entrenamiento, se compara el rendimiento de los modelos guardados, revisando las imágenes generadas (Figura \ref{fig:10}), comparando la calidad de la imagen de entrada, con el fin de observar qué tan bien delimita la nube de ceniza para elegir un modelo final.

Para la pérdida de la entropía cruzada y la precisión del modelo, se evalúa durante toda una época de interés, obteniendo las métricas (\textit{loss y accuracy}) tanto para el conjunto de entrenamiento como para el de validación. Una vez concluida la época, todos los valores de pérdidas y todos los valores de precisión son promediados para dar una probabilidad global o una puntuación del indicador, tanto para la sección de entrenamiento como validación.

Analizando la Figura \ref{fig:loss_acc} se determina que los mejores resultados se obtienen en la época 30 con una precisión del 60\% y pérdida de 37\%. Los valores obtenidos de la gráfica no determinan qué tan bueno es el modelo para la ejecución de la predicción ya que el modelo GAN no converge y su resultado se relaciona a la parte visual de qué tan bueno es el modelo para delimitar la nube de ceniza (Figura \ref{fig:10}).

\subsection{Matriz de Confusión}
Para conocer el rendimiento del modelo, aprovechamos la \textit{matriz de confusión} que nos permite analizar el número de predicciones correctas e incorrectas resumidas en una tabla de conteo que se muestra para cada clase (Tabla \ref{tab:confusion}). En esta tabla se representan los argumentos utilizados para determinar la matriz de confusión.

   \begin{table}[!htb]
    \centering
        \caption{Matriz de Confusión}\label{tab:confusion}
        \scalebox{0.7}{
        \begin{tabular}{|c|c|}
        \hline 
        \textbf{Verdadero Positivo} &  \textbf{Falso Positivo}\\
        \hline 
        Imagen satelital con nube de ceniza – Imagen blanco y negro con nube de ceniza & Imagen satelital sin nube de ceniza  – Imagen blanco y negro con nube de ceniza\\
        \hline
        \textbf{Falso Negativo} &  \textbf{Verdadero Negativo}\\
        \hline
        Imagen satelital con nube de ceniza – Imagen blanco y negro sin nube de ceniza & Imagen satelital sin nube de ceniza - Imagen blanco y negro sin nube de ceniza\\
        \hline
        \end{tabular}
        }
    \end{table}

\begin{figure}[!htb]
        \centering
        \includegraphics[width=\textwidth]{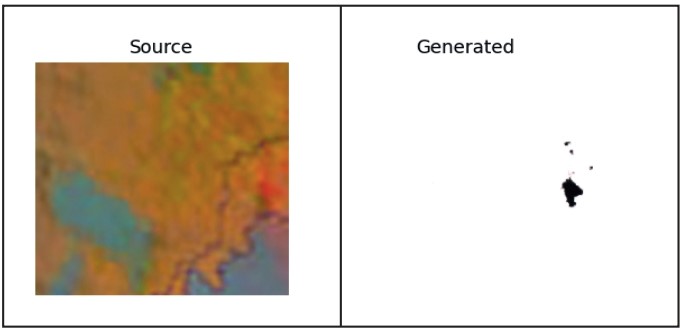} \caption{Ejemplo de la predicción del modelo en la época 90 utilizado para la elaboración de la matriz de confusión.}
        \label{fig:test}
    \end{figure}

Para la determinación de la matriz de confusión del modelo se utiliza la época 90 ya que presenta mejor la delimitación de la nube de ceniza. Consideramos el conjunto de datos del \textit{Test} que se encuentra en el sitio Web \textit{GitHub}\footnote{\url{https://github.com/GisseTorres04/GANs\_Pix2Pix\_Traduccion\_Imagenes\_Ceniza.git}} (\textit{Imágenes Test Matriz Confusión}) que comprende 30 pares de imágenes satelitales (Source) e imágenes en blanco y negro (Generated). Estas últimas se obtuvieron a partir de la predicción que se aplicó para el modelo de la época 90. Un ejemplo de las imágenes utilizadas para la determinación de la matriz de confusión aparece en la Figura \ref{fig:test}.

\begin{figure} [!htb]
        \centering
        \includegraphics[width=0.5\textwidth]{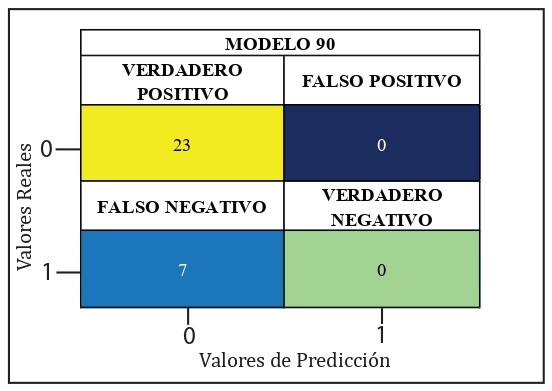} \caption{Matriz de confusión del modelo de la época 90 aplicado para la evaluación.}
        \label{fig:confusion}
    \end{figure}

Efectuamos el conteo manual de cada uno de los parámetros de la matriz de confusión de la Tabla \ref{tab:confusion}, obteniendo así los valores que reflejan cada uno de ellos, representado el verdadero positivo (cuadro amarillo) y verdadero negativo (cuadro verde) son los resultados donde el modelo predice correctamente si existe ceniza o no existe ceniza en la imagen, mientras el falso positivo y falso negativo (cuadros azul y celeste) son los resultados donde el modelo predice incorrectamente, obteniendo ceniza donde no existe o viceversa. Los resultados del conteo manual de la matriz de confusión se observan en la Figura \ref{fig:confusion}.

Los resultados de la matriz de confusión permiten hacer el cálculo de la \textit{exactitud} que representa el porcentaje de predicciones correctas frente al total \cite{TowardsConfussion}, obteniendo un valor de 0,7667, la \textit{precisión} es el número de elementos correctamente clasificados dividido entre el número total de elementos en el conjunto de pruebas \cite{Microsoft1}, obteniendo 1, la \textit{sensibilidad} y \textit{especificidad} son valores que nos indican la capacidad de nuestro estimador para discriminar los casos positivos de los negativos \cite{TowardsConfussion}, obteniendo como valores en la sensibilidad de 0,7667, mientras que la especificidad de 0 ya que no se presentan valores en los falsos positivos y verdaderos negativos para poder obtener su valor. Se determina que el desempeño del algoritmo es bueno, dando una evaluación positiva para la visualización de la delimitación de la nube ceniza, pero se debe tomar en cuenta que por ser una matriz de confusión realizada manualmente por medio de la visualización de los resultados obtenidos a partir de la predicción del dataset de test, se obtienen valores casi perfectos, depende de la observación de las imágenes que realice el humano, el cual puede ser subjetivo para la obtención de los valores.

\subsection{Predicción}

Para la predicción se utilizaron imágenes satelitales multiespectrales de nube de ceniza obtenidas del satélite GOES-16, mismas que están publicadas en la carpeta de \textit{Imágenes de Prueba} en \textit{GitHub}\footnote{\url{https://github.com/GisseTorres04/GANs\_Pix2Pix\_Traduccion\_Imagenes\_Ceniza.git}}. La predicción consiste en la carga de la imagen (\textit{load\_image}), la cual es redimensionada a 256x256 píxeles, posteriormente es convertida en matriz \textit{numpy} y los valores son escalados entre -1 y 1. Luego de realizar los cambios pertinentes de la imagen, se procede a cargar la imagen de origen ubicada en la carpeta mencionada, así como el modelo que está en la carpeta \textit{Modelos} con el formato \textit{h5}. Para finalizar, se procede a guardar la predicción de la imagen.

    \begin{figure} [!htb]
        \centering
        \includegraphics[width=\textwidth]{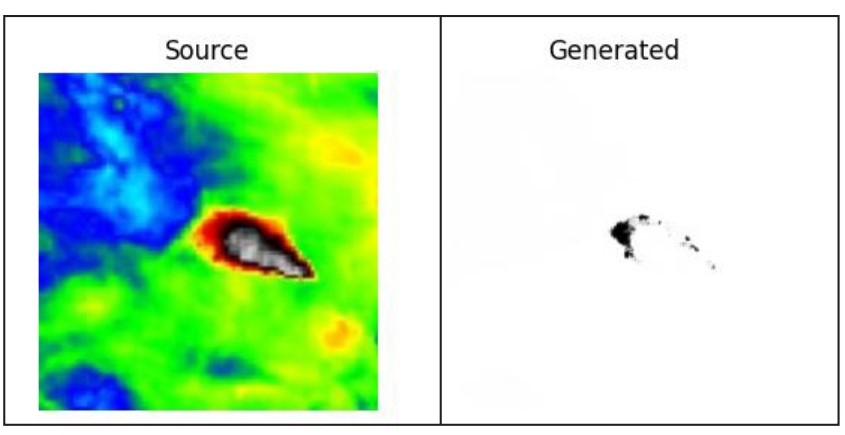} \caption{Predicción del Modelo. En el lado izquierdo una imagen satelital multiespectral GOES-16 del Volcán Sangay del Ecuador del 8 de febrero del 2022 - 09:40:30 UTC, mientras que la imagen del lado derecho, resulta de aplicar el modelo de la época 90, obteniendo la imagen en blanco y negro que representa la delimitación de la nube de ceniza.}
        \label{fig:prediccion}
    \end{figure}

El resultado de la predicción de ejemplo se presenta en la Figura \ref{fig:prediccion}, donde la parte izquierda corresponde a la imagen satelital de la nube de ceniza considerada como la imagen de entrada, mientras que la imagen de la derecha en blanco y negro representa la imagen de la delimitación de la nube de ceniza que identifica el contorno de la ceniza representada por el color negro. Consideramos una delimitación de ceniza aceptable ya que la imagen generada presenta el borde de la nube, dando a conocer la representación misma basada en la imagen satelital.

        \begin{figure} [!htb]
        \centering
        \includegraphics[width=\textwidth]{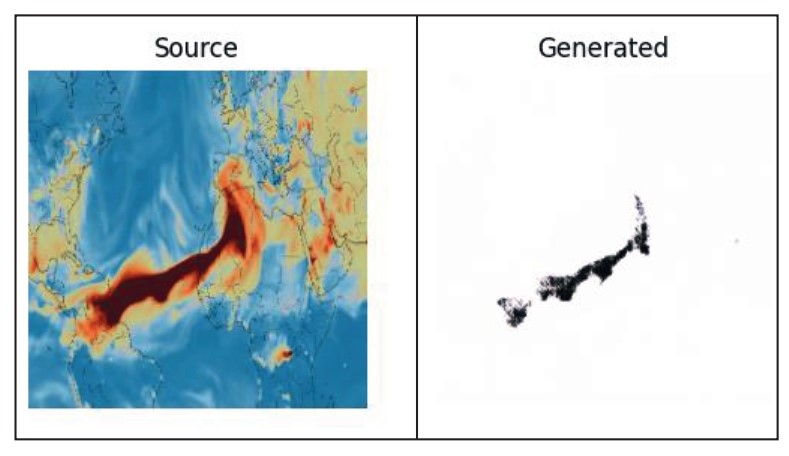}  \caption{Predicción del Modelo. En el lado izquierdo, la imagen satelital multiespectral GOES-16 del Volcán Cumbre Vieja, Isla de La Palma en las Islas Canarias, España, del 13 de abril del 2021, aplicando el modelo obtenido de la época 90.}
        \label{fig:prediccion2}
    \end{figure}

Otro ejemplo a partir del modelo Pix2Pix es la Figura \ref{fig:prediccion2}, donde la delimitación de ceniza se asemeja más a la imagen satelital. Comparando con la Figura \ref{fig:prediccion}, se determina que para una mejor delimitación, la imagen de entrada debe ser semejante a las que componen el dataset de entrenamiento, caracterizado por presentar colores distintos al blanco y negro, que sean colores llamativos (ejemplo color rojo) para que pueda detectar de mejor manera la delimitación de la nube de ceniza.

\section{Discusión}
\label{discusion}
Nuestro objetivo ha sido mejorar los resultados del trabajo de delimitación automática de ceniza volcánica efectuado en \cite{Aldas}, cuyo modelo está basado en redes neuronales convolucionales dispuestas en una arquitectura encoder-decoder (codificador - decodificador). Como alternativa, nuestro enfoque utiliza una red generativa adversaria, misma que se compone de un discriminador y un generador. La Tabla \ref{tab:comparacion} presenta una comparación entre los dos modelos, CNN y GAN, donde se pueden apreciar las características de los modelos con las respectivas ventajas y desventajas.

    \begin{table*}[!htb]
        \centering
        \caption{Comparación de los modelos CNN y GAN de la delimitación de nube de ceniza volcánica.}\label{tab:comparacion}
        \begin{tabular}{|m{3cm}|m{6cm}|m{6cm}|}

        \hline 
        \textbf{Fases} &  \textbf{CNN} & \textbf{GAN}\\
        \hline 
        Dataset &  El tratamiento de imágenes mediante el programa Online-Convert, transformando archivos GIF a JPG con un tamaño de 300x200 píxeles. & Descarga del dataset de imágenes satelitales e imágenes en blanco y negro de la delimitación de ceniza, organizadas en 2 carpetas, cada una con 600 imágenes de cada tipo.\\
        \hline
        Preprocesamiento & Construcción de las imágenes en blanco y negro con Photoshop CS6, redimensionamiento a un tamaño de 32x32 píxeles. Trazado del borde de la zona de nube de ceniza con el objetivo de facilitar el reconocimiento de la misma. & Redimensionamiento y unión de imágenes a través de Pinetools. En primera instancia, imágenes de tamaño 256x256 y seguidamente pares de imágenes combinadas (satelital e imagen en blanco y negro) con un tamaño de 256x512 píxeles.\\
        \hline
        División del dataset & Uso del comando train\_test\_split(), distribuyendo automáticamente los datos en dos subconjuntos train (80\%) y test (20\%). & División manual en base al tipo de satélite GOES 16 y 17, Meteosat-11 y Himawari-8. División en 3 subcarpetas: train (80\%), validación (15\%) y test (5\%).\\
        \hline
        Creación del modelo & Arquitectura CNN, declaración de los bloques codificador (encoder) y decodificador (decoder). & Arquitectura GAN, definición del modelo discriminador (red convolucional), generador (red neuronal U-Net), y GAN (generador-discriminador).\\
        \hline
        Entrenamiento & Uso de la plataforma Google Colab con GPU, el proceso tuvo una duración de 15 minutos para 100 épocas. & El entorno de desarrollo es Visual Studio Code con una configuración de CPU, el proceso duró 20 horas para 100 épocas.\\
        \hline
        Evaluación & Curvas de pérdida del entrenamiento y validación, curvas de precisión del entrenamiento y validación, y matriz de confusión. La mejor precisión del entrenamiento en la época 10 con un 99\% y una pérdida de 16\%. & Evaluación cada 10 épocas de entrenamiento, obteniendo 3 imágenes que representan las imágenes de entrada y la generada (Figura \ref{fig:10}). Como resultado, una pérdida del 30\% y una precisión del 60\% en base a las gráficas. La matriz de confusión es manual para la época 90, con una precisión de 1, satisfactoria para la delimitación de ceniza.\\
        \hline
        \end{tabular}
    \end{table*}

    \begin{figure} [!htb]
        \centering
        \includegraphics[width=\textwidth]{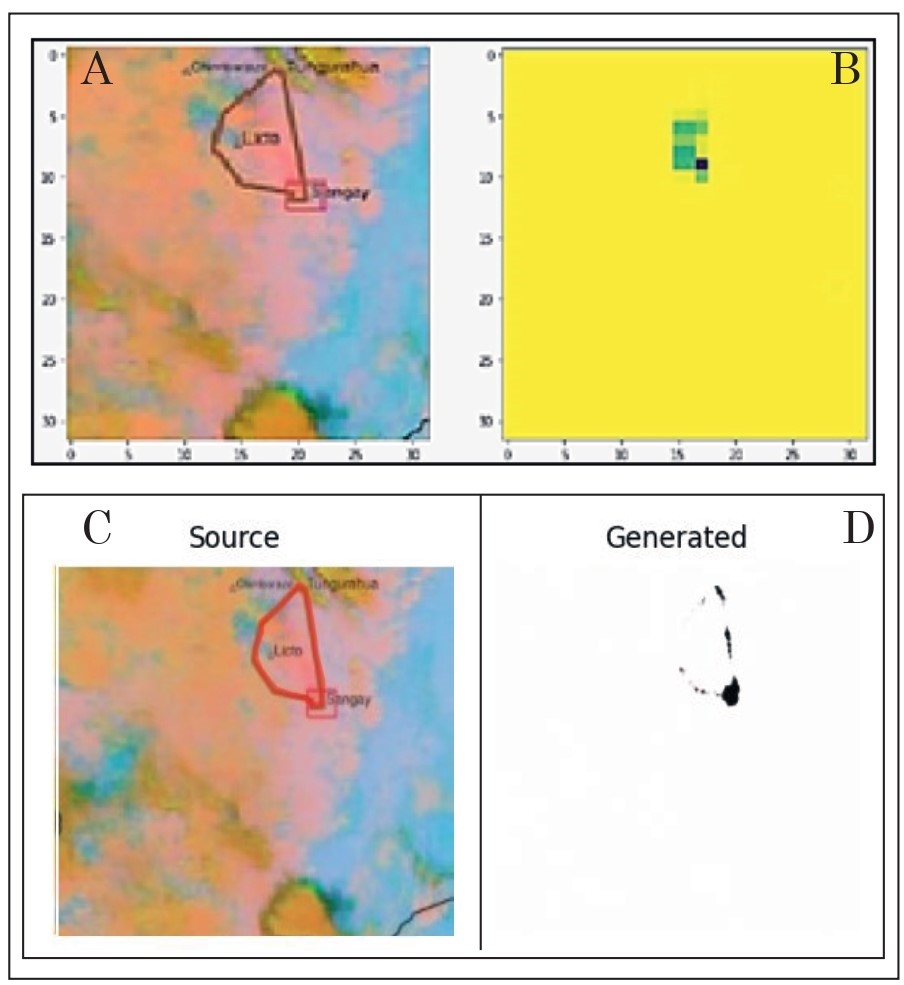} \caption{\textbf{A)} Imagen satelital multiespectral del Volcán Sangay obtenida del satélite GOES-16 el 9 de marzo de 2021; \textbf{B)} Imagen segmentada que delimita la nube de ceniza del Volcán Sangay generada por la CNN; \textbf{C)} Imagen satelital multiespectral del Volcán Sangay obtenida del satélite GOES-16 el 9 de marzo de 2021; y \textbf{D)} Imagen a blanco y negro que representa la delimitación de la nube de ceniza del Volcán Sangay generada por la red GAN.}
        \label{fig:results}
    \end{figure}

La Figura \ref{fig:results} incluye 4 secciones que representan la imagen satelital multiespectral (entrada), imagen blanco y negro e imagen amarillo y verde (salida), donde se observa que las imágenes resultantes (B y D) se diferencian al momento de delimitar la nube de ceniza, siendo el caso D más semejante a la imagen satelital, cuya delimitación de la nube de ceniza se acerca más al contorno de la misma. Esta imagen de predicción del modelo GAN comparada con la imagen del modelo CNN (B), presenta una mejor calidad, resolución, ajuste y detección de la delimitación de ceniza. Por tanto, la aplicación del modelo GAN con respecto al modelo CNN constituye una mejora en la traducción de imagen a imagen. Los resultados del modelo GAN son superiores a los obtenidos con el modelo CNN, esto se puede determinar con una inspección visual. El producto final es una imagen con buena resolución y detección de la delimitación de la nube de ceniza.

\section{Conclusiones y recomendaciones}
\label{conclusion}
La implementación utilizada está definida por Jason Brownlee en “Cómo desarrollar un Pix2Pix GAN para la traducción de Imagen a Imagen” en el sitio Web \textit{ Machine Learning Mastery}\footnote{\url{https://machinelearningmastery.com/how-to-develop-a-pix2pix-gan-for-image-to-image-translation/}}. El modelo GAN implementado para la delimitación de nube ceniza funciona de manera adecuada, los resultados obtenidos son superiores referentes a resolución, la calidad y delimitación de la nube de ceniza comparados con los resultados obtenidos a partir del modelo CNN. Sin embargo, una de las desventajas de aplicar el modelo Pix2Pix, son las características de las imágenes multiespectrales al entrenar el modelo, ya que al momento de predecir una imagen, si esta no presenta las características del dataset con el cual se entrenó el modelo, la predicción no es la mejor. Se determinó que mientras mayor es el volumen del dataset, se tendrá un mejor resultado en el mapeo de imágenes al momento de entrenar el modelo, así como la calidad de la imagen de salida está dada en función del tamaño de píxeles de las imágenes del dataset.

El modelo GAN puede ser utilizado de manera general para la delimitación  de ceniza de cualquier volcán del mundo, siempre y cuando la imagen de entrada contenga rasgos o características distintivas del dataset de entrenamiento, Estos rasgos o características son propias de las imágenes, en este caso al tratarse de imágenes satelitales multiespectrales corresponde a la gama de colores, tamaño de pixel, calidad de las imágenes de entrenamiento, etc.

Con base en las gráficas de pérdida, un valor promedio del 30\% en la época 30, comparada con la pérdida del modelo CNN con un valor menor a 16\%, no es un factor determinante, ya que por el hecho de presentar una pérdida menor, este no refleja una mejor calidad y resolución en la predicción. En el modelo CNN, se evalúan las pérdidas de manera general a comparación del modelo GAN, el cual analiza el comportamiento del modelo teniendo en cuenta que la pérdida del generador mejora cuando la pérdida del discriminador disminuye, es decir, se diferencian debido a que el modelo GAN no se relaciona con la pérdida o precisión durante el entrenamiento, sino se encuentra directamente relacionado con el ajuste visual de la imagen, dando una apariencia de la imagen a predecir con mayor calidad, resolución y delimitación de la nube de ceniza definiendo el mejor modelo, en nuestro caso, es el modelo de la época 90 que resalta las características mencionadas.

Las redes GANs, específicamente el modelo Pix2Pix, por su versatididad y adaptabilidad es una opción aplicable a la resolución de problemas en el campo de la geología. Entre las posibles variantes de aplicación se encuentran: la delimitación de deslizamientos a partir de imágenes satelitales de Google Earth, la identificación de minerales con base en fotografías, la identificación de estructuras geológicas a partir de fotografias de afloramiento, etc., por lo que el campo de aplicación es muy variado, siempre controlado por el dataset de imágenes que se utilice para entrenar el modelo y las imágenes deseadas como salida de la predicción. 

\printbibliography

\end{document}